\def\year{2022}\relax
\documentclass[letterpaper]{article} 
\usepackage{aaai22}  
\usepackage{times}  
\usepackage{helvet}  
\usepackage{courier}  
\usepackage[hyphens]{url}  
\usepackage{graphicx} 
\usepackage{natbib}  
\usepackage{caption} 
\DeclareCaptionStyle{ruled}{labelfont=normalfont,labelsep=colon,strut=off} 
\usepackage{amsfonts}       
\usepackage{nicefrac}       
\usepackage{microtype}      
\usepackage[table]{xcolor}
\usepackage{amsmath}
\usepackage{siunitx}
\usepackage{booktabs, makecell}
\usepackage{multirow}
\usepackage{subcaption}
\usepackage{algorithm,algorithmic}
\usepackage{xspace}
\usepackage{textcomp}

\definecolor{lightgray}{gray}{0.9}
\newtheorem{definition}{Definition}[section]

\newcommand{\D}{\mathcal{D}}
\newcommand{\Q}{\mathcal{Q}}
\newcommand{\N}{\mathcal{N}}

\newcommand{\binset}{\{0,\ 1\}}

\def \biased {shifted\xspace}

\makeatletter
\newcommand\para{\@startsection{paragraph}{4}{\parindent}%
	{-.25\baselineskip \@plus -.5\p@ \@minus -.1\p@}%
	{-3.5\p@}%
	{\ACM@NRadjust{\@parfont\@adddotafter}}}
\makeatother

\usepackage{newfloat}
\usepackage{listings}
\floatstyle{ruled}
\newfloat{listing}{tb}{lst}{}
\floatname{listing}{Listing}

\usepackage[capitalize]{cleveref}

\title{Black-Box Audits for Group Distribution Shifts}
\author{
    Marc Juarez\textsuperscript{\rm 1}\equalcontrib\footnote{M.\ Juarez did most of this research while at the University of Southern California.},
    Samuel Yeom\textsuperscript{\rm 2}\equalcontrib,
    Matt Fredrikson\textsuperscript{\rm 2}
}
\affiliations{
    \textsuperscript{\rm 1}University of Edinburgh\\
    marc.juarez@ed.ac.uk\\[0.1cm]
    \textsuperscript{\rm 2}Carneige Mellon\\
    {syeom,mfredrik}@cs.cmu.edu
  
}

\begin{document}

\maketitle

\begin{abstract}
When a model informs decisions about people, distribution shifts can create undue disparities.
However, it is hard for external entities to check for distribution shift, as the model and its training set are often proprietary.
In this paper, we introduce and study a black-box auditing method to detect cases of distribution shift that lead to a performance disparity of the model across demographic groups.
By extending techniques used in \emph{membership and property inference attacks}---which are designed to expose private information from learned models---we demonstrate that an external auditor can gain the information needed to identify these distribution shifts solely by querying the model.
Our experimental results on real-world datasets show that this approach is effective, achieving 80--100\% AUC-ROC in detecting shifts involving the underrepresentation of a demographic group in the training set. Researchers and investigative journalists can use our tools to perform non-collaborative audits of proprietary models and expose cases of underrepresentation in the training datasets.
\end{abstract}

\section{Introduction}\label{sec:intro}
With the widespread adoption of machine learning (ML), developers of ML models face new practical challenges.
One such challenge is \emph{distribution shift} (DS), also known as \emph{dataset shift}, which occurs when a model's training data distribution differs from that encountered during deployment.
Because this distributional mismatch can lower the model's accuracy in practice, much research in ML has been devoted to detecting DS~\cite{shafaei2018less}.
However, less work has focused specifically on DSes that cause a disparate performance of the model across demographic groups.\looseness=-1


%
The disparate performance of a model across demographic groups is a concern in the context of fair ML.
ML models are increasingly used to make consequential decisions about people, such as college admissions, loan approvals, and hiring.
A DS in the form of \emph{underrepresentation}, where the training set of the model is not representative of the population, may cause a performance disparity to the detriment of the underrepresented groups.
For example, K\"{a}rkk\"{a}inen and Joo~\cite{karkkainen2019fairface} show that racial minorities are underrepresented in public benchmark face datasets, and demonstrate a stark accuracy disparity across racial groups in state-of-the-art face recognition models, pointing at the underrepresentation as the reason for the disparity.
As these models can have a broad societal impact, investigative journalists~\cite{larson2017how} and researchers~\cite{buolamwini2018gender} often perform external audits to uncover biases in proprietary models.

In this paper, we aim to facilitate such audits by considering an adversarial setting where the deployment of a model is audited for underrepresentation without the cooperation of the model holder.
We present a \emph{black-box} auditing method, which requires only query access to the audited model and some information about the learning algorithm.
As companies often sell query access to their proprietary models, our audit can be applied to such models even when the model holder is indifferent or hostile to the auditing effort.
Moreover, we argue that the model holder cannot reliably detect an audit in progress and adjust its response accordingly (see \cref{sec:discussion}).
Therefore, our auditing method can be leveraged to encourage a reluctant model holder to address DS and its harms.\looseness=-1

For example, even if a model is much less accurate on a racial minority, it is not clear how to rectify this, as the choice of learning algorithm~\cite{hooker2021moving} or features~\cite{chen2018why} can also cause an accuracy disparity.
On the other hand, if our audit reveals an unrepresentative training set, the model holder can be more effectively pressured into obtaining a more balanced training set.

Our technical contributions are as follows:

\begin{itemize}
	\item We formally define the distribution shift audit and characterize it as a game between the auditor and the entity holding the model (\cref{sec:statement}).
	\item We propose a novel black-box DS auditing technique that can detect underrepresentation issues in the training distribution of a model (\cref{sec:auditmethod}). The technique is based on a theoretical analysis that shows how model overfitting can leak information about the model's disparate behavior across groups (\cref{sec:fairnessaudit}).
	\item We evaluate our auditing technique to quantify its performance in several practical scenarios.
        \cref{sec:experimentmethod} describes the evaluation methodology and \cref{sec:eval} presents its results.
        Our auditing techniques achieve 80--100\% AUC-ROC for detecting a shift of underrepresentation.
\end{itemize}

\section{Problem Statement}\label{sec:statement}
In this section, we formally define the audit and the DS that it aims to detect.

%

\subsection{The Distribution Shift Audit}
To formally define the auditor's task in detecting a DS, we propose a game definition based on the one that prior work has proposed for membership inference~\cite{yeom2020overfitting}.
In our new definition, the auditor is given query access to a model that was either trained on $\D$ or $\D'$ depending on the outcome of a secret bit $b$ sampled uniformly at random.
$\D$ represents the \emph{normative} distribution that the model should be trained on, and $\D'$ is an alternative distribution that the model could be trained on instead of $\D$.
The auditor wins the game if she guesses $b$ correctly.
\begin{definition}[DS audit]
	\label{def:midistauditgame}
	\setlength\itemsep{0.05em}
	Let $A$ be a learning algorithm, $n$ be a positive integer, and $\D$ and $\D'$ be distributions over data points $(x,\ y)$. The DS audit game proceeds as follows:
	\begin{enumerate}
		\item The challenger chooses $b \gets \binset$ uniformly at random.
		\item The challenger samples the training set $S \sim \D^n$ if $b = 0$, and $S \sim (\D')^n$ if $b = 1$.
		\item The challenger trains the model $h_S = A(S)$.
		\item The auditor is given black-box access to $h_S$. The auditor also knows $A$, $n$, and $\D$, which are assumed to be public information.
	\item The auditor wins if she outputs the correct value of $b$.
	\end{enumerate}
\end{definition}

\paragraph{Threat model.}
As stated in \cref{def:midistauditgame}, we assume a black-box audit throughout this paper, i.e., the auditor can query the audited model and observe its output on any chosen input, but she cannot observe the model parameters and intermediate computations.
This setting is a more realistic model of the capabilities of an external auditor than the white-box model considered in prior work~\cite{ganju2018property,ateniese2015hacking}, but it is also more technically challenging.
In addition, the auditor does not have access to the model's training dataset and cannot tamper with it or the data collection process. The auditor only knows the size of the training set, $n$, the learning algorithm used to train the model, $A$, and its hyperparameters.
An auditor can infer these details from the type of learning task, patents, white-papers, other documentation published by the company that produced the model~\cite{blooface2020,syndicai2020,pragli2020,clarifai2020}, and through other black-box techniques~\cite{oh2019towards}.
We challenge the assumption that the adversary knows $n$ and show that this assumption can be relaxed in practice (see \cref{sec:available_data}).

\looseness=-1
Finally, the normative distribution $\D$ is public, but the auditor does not know $\D'$.
The audited entity also knows $\D$ but may end up training on $\D'$ for various reasons.
For instance, $\D$ could be the distribution of the US population and $\D'$ the distribution in a few locations selected for data collection.

\subsection{The Group Distribution Shift Audit}

%
The main motivation for our audit is to detect a specific type of DS: a shift caused by underrepresentation, wherein a demographic group is not as prevalent in the model's training set as it \emph{should} be.
\cref{def:groupshift} formalizes this type of DS, which we name \emph{group distribution shift} (GDS).
Here, the random variable $Z$ denotes the relevant demographic attribute, and $X$ and $Y$ are the input features and the response, respectively.

\begin{definition}[Group distribution shift]
	\label{def:groupshift}
	The alternative distribution $\D'$ is group-shifted from $\D$ if
	{\small $$\Pr_{(x,y,z) \sim \D}[X{=}x, Y{=}y \mid Z{=}z] \;= \Pr_{(x,y,z) \sim \D'}[X{=}x, Y{=}y \mid Z{=}z]$$}
	for all possible values of $x$, $y$, and $z$, but the marginal distribution of $Z$ is different for $\D$ and $\D'$, i.e., there exists $z$ such that\looseness-1
	$$\Pr_{(x,y,z) \sim \D}[Z{=}z] \;\neq \Pr_{(x,y,z) \sim \D'}[Z{=}z].$$
\end{definition}

A possible cause for GDS is a sampling bias.
Consider a loan default prediction model, where the data collection overlooked poor neighborhoods.
In the US, poor neighborhoods disproportionately consist of racial minorities, so this data collection practice would lead to a racial sampling bias.
Although the bias does not necessarily imply an unfair model, it may contribute to an inter-group accuracy disparity where other types of DS would not, so the auditor should be able to detect GDS specifically.

Thus, the audit that we consider is a \emph{GDS audit}, a DS audit where the shift between $\D$ and $\D'$ in \cref{def:midistauditgame} is a GDS. Throughout the paper, we assume that $\D$ is the \emph{normative} distribution (e.g., a group-balanced distribution), and $Z \in \binset$, where $Z=1$ encodes the underrepresented group.\looseness-1

\section{The Auditing Techniques} \label{sec:auditmethod}
In this section, we describe the techniques used to implement our audits.
\cref{sec:mishadow} provides an overview of the shadow model setup as used in membership and property inference, and how it compares to our setup. In \cref{sec:fairnessaudit}, we provide a theoretical analysis that informs the design of our GDS audit procedure. Finally, \cref{sec:attacktype} describes the components of the GDS audit procedure.

\subsection{Shadow Training} \label{sec:mishadow}
The attacker's goal in membership inference is to distinguish points of the target model's training set $S$ versus points of the general population.
However, since the attacker cannot access $S$, he does not have examples of the target model's behavior on it.
Shokri et al.~\cite{shokri2017membership} introduced \emph{shadow training} to procure such examples.
By sampling a new set $S_i$, the attacker can use the same learning algorithm, to train a (shadow) model $h_{S_i}$ that imitates the target model's behavior if it had been trained on $S_i$.
Then, the attacker feeds the outputs of $h_{S_i}$ to a meta-classifier, the \emph{attack} model, that learns to distinguish members of the shadow model's training from its outputs.

However, the attacker's ultimate goal is to identify the training set of the target model, not of the shadow model.
Therefore, the attacker has the attack model learn from shadow models trained on many different sets.
This makes it more likely that the attack model learns patterns specific to the learning algorithm $A$ and generalizes to the target model.

Similarly, property inference also uses shadow training to determine a property of the training set.
By training multiple shadow models on datasets with and without the property, the attack model learns to identify whether the target model's training set satisfies the property.
By contrast to membership inference, existing property inference attacks~\cite{ateniese2015hacking,ganju2018property} train the attack model with the parameters of the shadow models~\cite{ateniese2015hacking,ganju2018property}, which implicitly assumes white-box access to the target model at testing time.

\paragraph{Black-box shadow training for auditing.}
Our GDS audit applies shadow training similarly to property inference: it uses samples from two different distributions to train the shadow models and the attack model learns to distinguish whether the target model (the audited model) was trained on data from $\D$.
However, because the auditor only has black-box access to the audited model, the audit's attack model is trained only with the input-output behavior of the models instead of their parameters.

\begin{figure}
  \centering
	\includegraphics[scale=0.58]{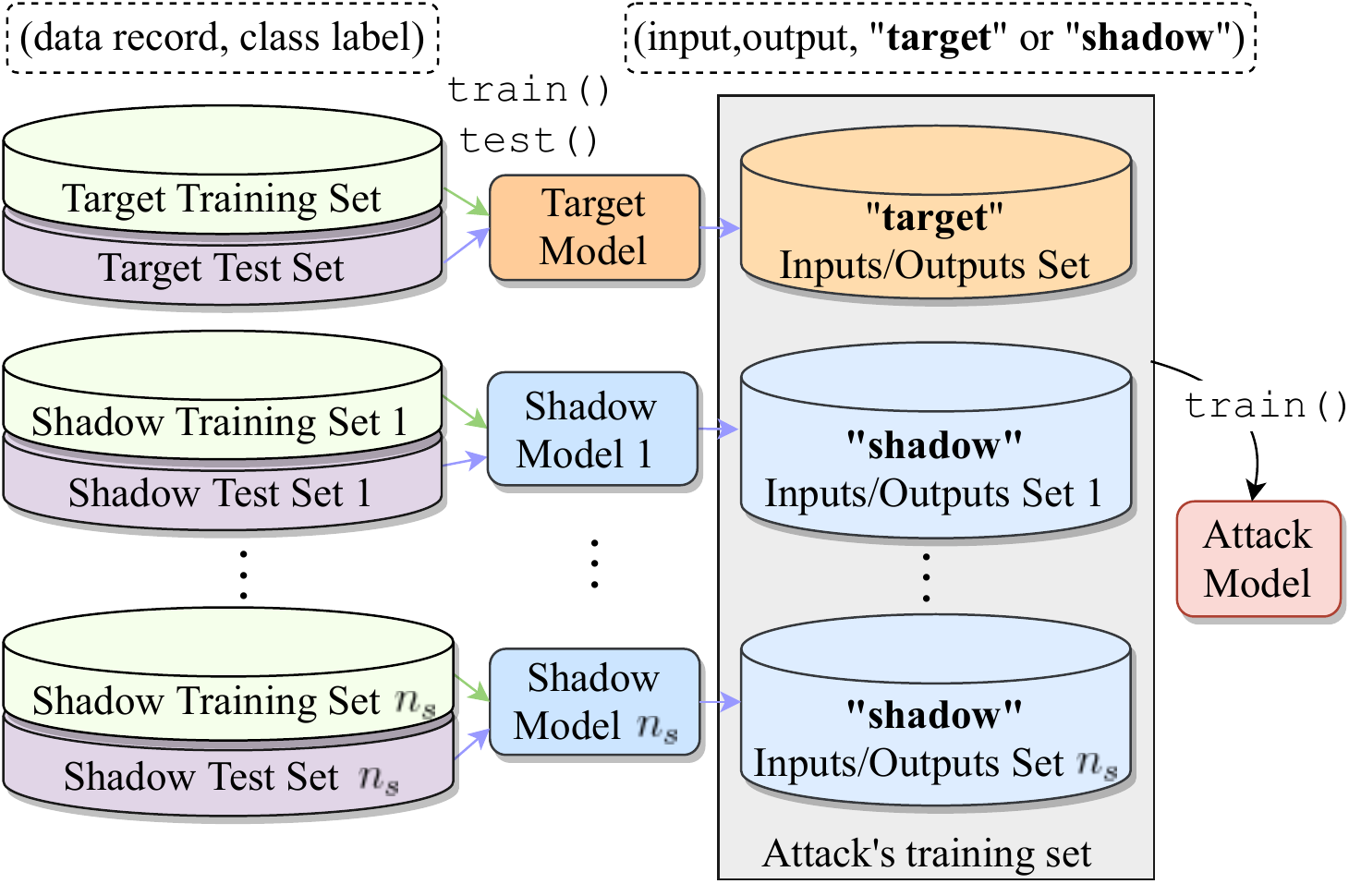}
	\caption{A schematic for training the audit's attack model. To obtain the attack model's training set, the auditor trains and tests the target and shadow models. Then, their inputs and outputs on a test set are labeled with whether the model was the target or a shadow model.}
  \label{fig:shadowaudits}
\end{figure}

\cref{fig:shadowaudits} depicts this setup---the attack model's training set consists of a set of input points and the corresponding outputs from the target and shadow models.
Each input/output pair is then labeled with a bit indicating whether it came from the target or a shadow model.
Because a single input/output pair may not contain enough information, the attack model is trained with $n_q$ pairs at a time, all generated from the same model.
Finally, at test time, the attack model is given $n_t$ pairs at a time and must guess either ``target'' or ``shadow''---$n_t$ (at testing) can be much larger than $n_q$ (at training).

Note that a trivial attack model can attain 50\% accuracy.
Moreover, if $b = 0$, i.e., $S$ is drawn from $\D$, the target model's behavior will be similar to those of the shadow models, so it will be hard for an attack to attain an accuracy significantly greater than 50\%.
Therefore, if the attack wins significantly more than half of the time, the auditor may conclude that $S$ was drawn from $\D'$.
However, although this approach may detect a general DS, in the next section we show it is not sufficient to specifically detect a GDS, and give an heuristic argument for why the attack's \emph{inter-group difference} in accuracy is better suited for detecting a GDS than its overall accuracy.

\subsection{The Auditor's Advantage} \label{sec:fairnessaudit}

We now theoretically analyze the auditor's success in a GDS audit when she uses black-box shadow training.
From this analysis, we will conclude that, at least for learning algorithms that tend to overfit, GDS can be detected by measuring the attack model's \emph{inter-group accuracy}, i.e., the difference in accuracy between groups.

Consider a 1-dimensional input $X \in \mathbb{R}$, binary $Y$ and $Z$.
The distribution of the $Z=1$ group, $\D_1$, satisfies $\Pr[Y{=}0 \mid Z{=}1] = \Pr[Y{=}1 \mid Z{=}1] = \frac{1}{2}$, with $X \mid Y{=}0, Z{=}1 \sim \N(-1+\tau, 1)$ and $X \mid Y{=}1, Z{=}1 \sim \N(1+\tau, 1)$.

For the $Z=0$ group, the distribution $\D_0$ is defined similarly, but $\tau$ is always zero.
The parameter $\tau$ is allowed to vary for $\D_1$ and indicates the degree to which the two groups differ.
Finally, the population distribution $\D$ is a 50-50 mix of $\D_0$ and $\D_1$.

We now train the target model $h_t$ and shadow model $h_s$.
To simulate an unrepresentative training set, we will assume that all points in the target model's training set $S_t$ are drawn from $\D_0$, whereas the shadow model is trained from a balanced training set $S_s$ drawn from $\D$.
The attack model is given either $h_t$ or $h_s$, each with probability $\frac{1}{2}$, and it queries the model on a random point $x$ to determine which one was given.\looseness-1

We model overfitting by assuming that the model is correct with higher probability when the query point $x$ is close to a point in the training set $S$, i.e., a model $h$ overfits if for an $\epsilon > 0$ we have
\[
\Pr[h(x) = y] =
\begin{cases}
\pi_\mathrm{tr}, & \text{if } \exists x' \in S \text{ s.t. } \|x - x'\| \le \epsilon \\
\pi_\mathrm{te}, & \text{otherwise.}
\end{cases}
\]
where $y$ is the correct response, and $\pi_\mathrm{tr}$ and $\pi_\mathrm{te}$ are probabilities of being correct with $\pi_\mathrm{tr}> \pi_\mathrm{te}$.

Now, our theoretical attack model uses only information about the queried model's accuracy.
In particular, without loss of generality, the attack model guesses ``target'' if the model $h$ is correct on the queried point $x$, and ``shadow'' otherwise.
For brevity, let $f_t(\Q) = \Pr_{x \sim \Q}[\exists x' \in S_t \text{ s.t. } \|x - x'\| \le \epsilon]$ and we define $f_s(\Q)$ analogously. 
Then, when querying a point $x \sim \Q$, the attack model's accuracy is
\begin{align*}
Acc \;=\; & \textstyle\frac{1}{2} \, \big(\Pr[\text{attack guesses } h_t \mid \text{it was given } h_t] \;\;+\\ &\quad\quad\; \Pr[\text{attack guesses } h_s \mid \text{it was given } h_s]\big) \\
\;=\; &\textstyle\frac{1}{2} \, \big(\Pr_{x \sim \Q}[h_t(x) = y] \;\;+\;\; \Pr_{x \sim \Q}[h_s(x) \neq y]\big) \\
\;=\; & \textstyle\frac{1}{2}\, \big(\pi_\mathrm{tr} \,f_t(\Q) + \pi_\mathrm{te} (1 - f_t(\Q)) \;\;+ \\&\quad\quad\;(1 - \pi_\mathrm{tr}) f_s(\Q) + (1 - \pi_\mathrm{te}) (1 - f_s(\Q))\big) \\
\;=\; & \textstyle\frac{1}{2}\;+ \;\frac{1}{2}(\pi_\mathrm{tr} - \pi_\mathrm{te}) (f_t(\Q) - f_s(\Q)).
\end{align*}
This means that the auditor's advantage over random guessing is proportional to $f_t(\Q) - f_s(\Q)$.

\begin{figure}
\centering
	\includegraphics[scale=0.9]{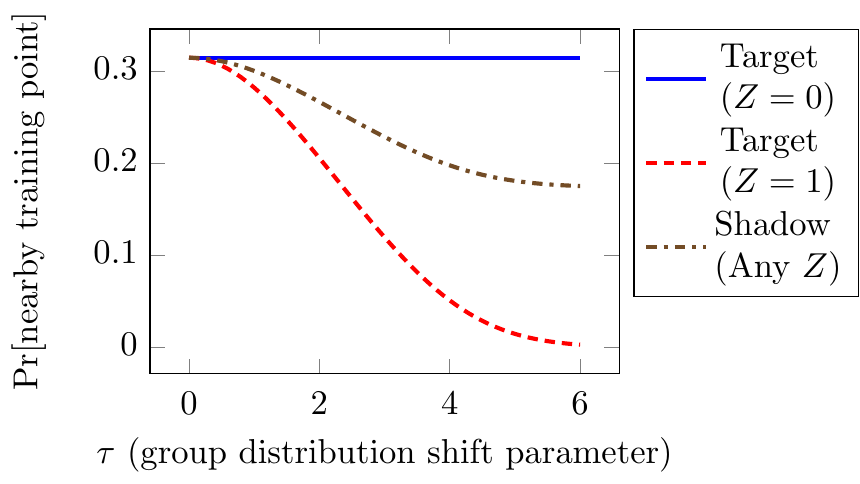}
	\caption{
		Probability that a random data point is close to a training set point for the models analyzed in \cref{sec:fairnessaudit}.
		The target model has an unrepresentative training set, and the shadow model has a demographically balanced training set.
		If the models overfit ($\pi_\mathrm{tr} > \pi_\mathrm{te}$), this difference in probability can allow an auditor to detect a GDS.
	}
	\label{fig:theoryplot}
\end{figure}

\cref{fig:theoryplot} plots the values of $f_t(\Q)$ and $f_s(\Q)$ for the case where $\epsilon=0.001$ and the training set consists of a total of 1,000 i.i.d.\ points.
Although we choose these specific values for concreteness, the patterns shown here generalize to many different values of $\epsilon$ and training set sizes.
The solid and dashed lines correspond to $f_t(\D_0)$ and $f_t(\D_1)$, respectively, and the dash-dotted line represents $f_s(\D) = f_s(\D_0) = f_s(\D_1)$.

Since $\D$ is a 50-50 mix of $\D_0$ and $\D_1$, $f_t(\D)$ is simply the average of the solid and the dashed lines.
This average is close to the dash-dotted line, so the quantity $f_t(\D) - f_s(\D)$ is close to zero, which means that the attack model is not much more accurate than random guessing.
However, if we consider the two protected groups separately, for sufficiently large GDS parameter $\tau$, the quantity $f_t(\D_0) - f_s(\D_0)$ is a large positive number, and $f_t(\D_1) - f_s(\D_1)$ is a large negative number.
This means that, if the learning algorithm overfits, i.e., $\pi_\mathrm{tr} - \pi_\mathrm{te}$ is large, the inter-group difference in the attack model's performance will also be large.

The above analysis motivates the attack model of our GDS audit. The shadow and attack model setup in both the control and \biased settings proceeds identically to that described in \cref{sec:mishadow}.
However, instead of measuring the attack model's accuracy, the auditor measures how much more accurate the attack model is on points drawn from $\D_0$ compared to $\D_1$. 

\subsection{The GDS Audit Procedure}\label{sec:attacktype}
In this section, we specify the concrete attack model that we propose for the GDS audits. Because of the variance in the outcomes of the $n_t$ queries to the attack model, the audit is implemented as a controlled experiment and the final audit decision as a statistical test.


\paragraph{The attack model.}


Recall from \cref{def:groupshift} that GDS occurs when a demographic group is underrepresented in the alternative distribution $\D'$ compared to the normative distribution $\D$.
Following the heuristic argument in \cref{sec:fairnessaudit}, the auditor will use the inter-group difference in the performance of the attack model to decide whether GDS has occurred.
Thus, there is no evidence of GDS if the attack model performs well on all groups, but if it performs significantly better on one group, the auditor will suspect GDS even if the overall performance is lower than random guessing.

In the theoretical analysis, the attack model uses only the accuracies of the target or shadow model to determine the queried model. The audit's attack model follows this approach and describes the input-output behavior of a model as its average performance (accuracy for classification; mean squared error for regression) on $n_q$ queries.
In our implementation, we train the attack model using logistic regression and then use it to predict whether the queried model is a target or a shadow model.
We have experimented with a more complex attack that also takes the query inputs (see \cref{sec:appendix_complex}), but discarded it because it compared unfavorably with the attack that only takes average performances.
To allow a direct comparison between attack's training and testing samples, we always let $n_q = n_t$.



\paragraph{The controlled experiment.}
Even when $b = 0$, the random draws of the training set from $\D$ may result in target models significantly different from the shadow models, especially if $A$ tends to overfit. As a result, there is variance in the attack model's measurements of the inter-group accuracy.
To address this issue, we take a hypothesis testing approach with the following null hypothesis ($H_0$): \emph{the audited model's training distribution is the normative distribution $\D$}.

To test $H_0$, the auditor performs a controlled experiment, which is divided into two settings:

\setlength{\leftmargini}{25pt}
\begin{itemize}

	\item The \emph{control} setting: Following the attack model setup described above, the auditor draws data from $\D$ to train $n_s$ shadow models, and a \emph{target model} which tries to mimic the audited model as if it had been trained on $\D$.\looseness-1
	
	\item The \emph{\biased} setting: The auditor follows the same steps as in the control setting, but she does not train a model for the target model---the target model now becomes the audited model.
\end{itemize}

\begin{figure}
	\includegraphics[scale=0.55]{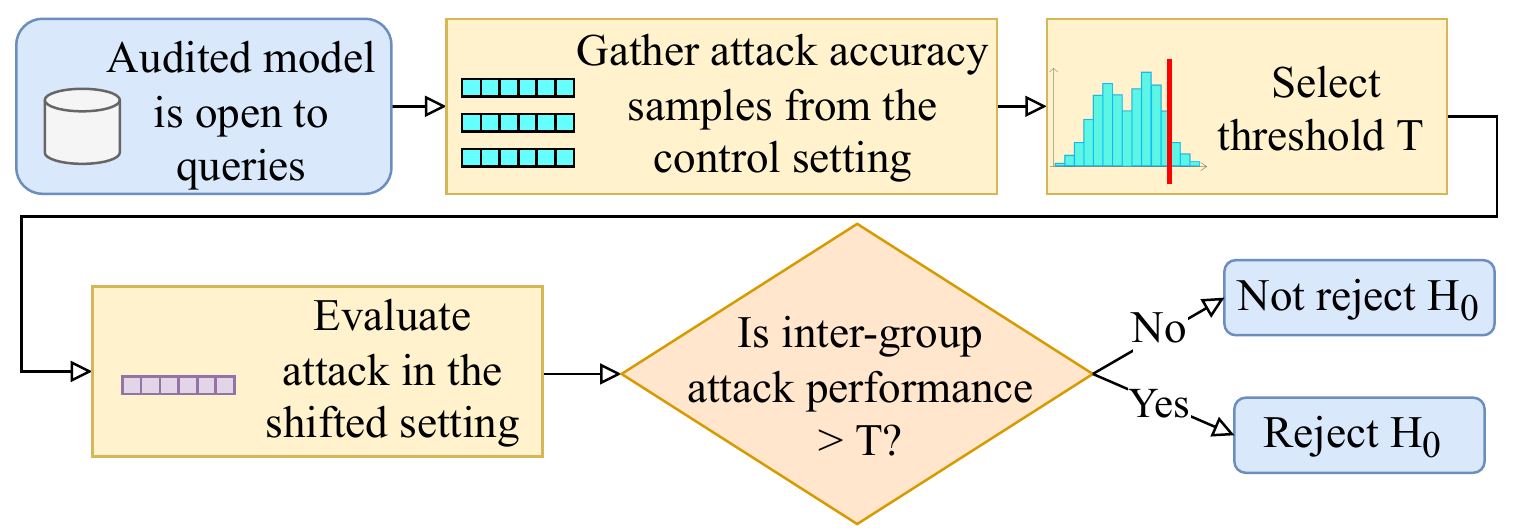}
	\caption{Diagram of the audit's workflow. The auditor has access to the public distribution $\D$ and uses it to train the models in the control and \biased settings. The audited model (whose training is possibly drawn from $\D'$) is queried in the \biased setting only. The threshold is taken from the attack model's accuracy distribution in the control setting and applied to the attack model's accuracy in the \biased setting.}
	\label{fig:audit}
\end{figure}

In both settings, the auditor trains and evaluates the \emph{attack} model (see \cref{fig:audit}).
The control setting evaluates the attack's inter-group performance when the target and shadow models' training sets are drawn from the same distribution.
By repeating the experiment in the control setting multiple times, the auditor can measure the extent to which the attack's inter-group performance can be attributed to the randomness in training (e.g., the specific training dataset).
In the \biased setting, the auditor is given only one model to audit, so she can only make one measurement of inter-group performance.

As in the original membership inference attack, the auditor could train multiple target models in the control setting, and multiple shadow models in both control and \biased settings to reduce the noise in her estimate of the attack's inter-group performance. However, in our evaluation we do not observe a significant increase in the audit's success by training multiple shadow models. In the rest of the paper, we always let $n_s=1$.

\paragraph{The statistical test.} To determine whether the difference between control and \biased inter-group performance of the attack is significant, the auditor performs a statistical test.
First, she determines a threshold at which she would reject $H_0$ from her observations in the control setting.
In \cref{sec:measurement}, we describe how we choose the threshold for our empirical evaluation, but this is not intended to be prescriptive---the threshold choice depends on the specific use case and the relative cost of false positives versus false negatives.
After setting the threshold, the auditor measures the attack's inter-group performance in the \biased setting.
If the measured inter-group performance is above the threshold, the auditor can be reasonably confident that $S$ was not drawn from $\D$.


\section{Evaluation Methodology} \label{sec:experimentmethod}
In our experiments, we introduce a shift between $\D$ and $\D'$ to simulate the audit and measure its performance. We now detail procedures common to all experiments and the datasets we used.

\subsection{Datasets}

\paragraph{US Census datasets.} We use the \emph{folktables}~\cite{ding2021retiring} Python module to access the US Census.
The \emph{folktables} project intends to replace UCI Adult as the golden dataset to benchmark fair ML algorithms and evaluate DSes on demographic data. We use the predefined California 2018 dataset with the income prediction task (\texttt{ACSInc}).
To artificially simulate a shift along the sex attribute, we partition the data into a dataset that follows California's sex distribution and one with only males. 

\paragraph{Face datasets.} We use two popular datasets for training face recognition models: \texttt{UTKFace}~\cite{utkface} and \texttt{CelebA}~\cite{celeba}. The main difference between the two datasets is that \texttt{CelebA} is composed by faces of celebrities.
Our target and shadow models predict a person's gender when given their face.

\paragraph{Medical dataset.} The \texttt{Warfar}~\cite{iwpc} dataset contains medical, genetic, and demographic features about patients who were prescribed Warfarin. The task is to predict the Warfarin dosage of a patient.

\medskip
In \cref{sec:appendix_arch}, we detail the implementations of the target and shadow models (including the audited model) that solve the tasks defined by these datasets.

These datasets represent use cases where it is desirable that the model be equally accurate on all subgroups of the population.
When this property does not hold, the auditor should ideally be able to point to a cause of the disparity so that specific remedies can be taken.
Here, the auditor specifically tests for an unrepresentative training set as a potential cause.
For example, since warfarin doses depend on the patient's race~\cite{iwpc}, it is important that the training sets of such model be racially diverse. 
Therefore, unlike in the general DS setting, we consider a DS to be an issue if, and only if, it changes the marginal distribution of the protected attribute (e.g., race, gender, or age).

\subsection{Measurements} \label{sec:measurement}
For both the control and \biased settings, we partition our data into five subsets---training sets for the target, shadow, and attack models, a shared test set for target and shadow models, and an attack model test set.
In the control setting, all sets are drawn from $\D$, whereas in the \biased setting, the target model training set is drawn from $\D'$.
Although the training set of the audited model may be drawn from either $\D$ or $\D'$ depending on the value of $b$, we always draw from $\D'$ in the \biased setting, as the audit's performance when $b = 0$ can be measured through the control setting. 

We run each setting multiple times with different random seeds, and the partitions change depending on the seed.
We indicate the exact number of seeds in \cref{sec:appendix_arch}.
Note that the auditor would only be given one model to audit during the deployment of the audit.
However, to estimate the test's statistical power (true positive rate), we also repeat the \biased setting experiment multiple times.
The statistical power indicates how likely the auditor is to correctly reject $H_0$.\looseness-1


We calculate the true positive rate by setting the auditor's threshold at the 90th percentile of the control values, which corresponds to a 10\% false positive rate.
We apply this non-parametric approach, rather than a $t$-test as done by Maini et al.~\cite{maini2020dataset}, because here we cannot assume that the control performance follows any specific distribution.
To measure how effective the audit is with other thresholds, we also report the area under the receiver operating characteristic (AUC-ROC) curve.


\section{Evaluation}\label{sec:eval}
We now describe our experiments in detail and present the results.
Our attack is successful in detecting GDS, with decreasing statistical power for smaller effect sizes.
Also, our audit significantly overperforms a naive audit that flags underrepresentation when there is an accuracy disparity.
Finally, we demonstrate that a false positive GDS audit is unlikely in the presence of DSes that are not a GDS.

\begin{table*}[t]
	\centering
	\caption{Configurations of the experiments and the results of the GDS audits. The split size ($m$) is the size of the data partitions. $\D_0$ and $\D_1$ are the over- and underrepresented groups, respectively. We report the target model's performance (accuracy on all rows except for the drug dosing task, which is the mean squared error) and standard  deviation in the \biased setting for the two protected groups.
	This is followed by the audit's TPR($P_{90\%}$) and AUC for the baseline (Base) and the attack (Atk).
	}
	\resizebox{\textwidth}{!}{
	\hskip-.3cm
        \begin{tabular}{ccccccccccccc}
        \toprule
        \multicolumn{7}{c}{\textbf{Experiment Configuration}} & 
        \multicolumn{2}{c}{\textbf{Target Model Perf.}} & 
        \multicolumn{2}{c}{\textbf{AUC}} &
        \multicolumn{2}{c}{\makecell{\textbf{TPR} ($\boldsymbol{P_{90\%}}$)}}\\\cmidrule(lr){1-7}\cmidrule(lr){8-9}\cmidrule(lr){10-11}\cmidrule(lr){12-13}
            \textbf{Task} & \textbf{Dataset} & \textbf{$\boldsymbol{\beta}$} & $\boldsymbol{\D_0}$ & $\boldsymbol{\D_1}$ & $\boldsymbol{m}$ & 
            $\boldsymbol{A}$ &
            $\boldsymbol{\D_0}$ &
            $\boldsymbol{\D_1}$ &
            \textbf{Base} & \textbf{Atk} &
            \textbf{Base} & \textbf{Atk} \\
                    
        			\cmidrule(lr){1-13}
        			
        			\multirow{2.5}{*}{\makecell{Gender \\ recog.}} & \texttt{UTKFace} & 1.0 &  White & \makecell{non-White} & 3.4K 
        			& CNN & $85.04\pm1.56$\% & $79.05\pm2.18$\% &
        			99.56\% & 79.68\% & 99\% & 64\%  \\
        			
        			\addlinespace
                                     
        			& \texttt{CelebA} & 1.0 & young & old & 18.4K 
        			& CNN & $95.36\pm0.45$\% & $92.43\pm1.41$\% &
        			99.62\% & 99.96\% & 99\% & 100\%  \\
        			\cmidrule(lr){1-13}
        			\multirow{5}{*}{\makecell{Drug \\ dosing}} & \multirow{5}{*}{\texttt{Warfar}} & 1.0 & \multirow{5}{*}{White} & \multirow{5}{*}{Asian} & \multirow{5}{*}{600} 
        			& \multirow{5}{*}{MLP} & $1.15\pm0.22$ & $6.77\pm1.49$ &
        			100\% & 100\% & 100\% & 100\%  \\
        			
        			& & 0.9 & & & & & 
        			$1.19\pm0.22$ & $1.52\pm0.28$ &
        			84.76\% & 91.74\% & 63\% & 82\%  \\
        			
        			& & 0.8 & & & & & 
        			$1.21\pm0.22$ & $1.12\pm0.16$ &
        			40.84\% & 62.86\% & 17\% & 38\%  \\
        			
        			& & 0.7 & & & & & 
        			$1.23\pm0.22$ & $1.00\pm0.14$ &
        			13.48\% & 57.80\% & 0\% & 24\%  \\
        			
        			& & 0.6 & & & & & 
        			$1.27\pm0.22$ & $0.95\pm0.12$ &
        			5.00\% & 49.40\% & 0\%  & 12\%  \\
        			\cmidrule(lr){1-13}
        			\multirow{5}{*}{\makecell{Income \\ Pred.}} & \multirow{5}{*}{\texttt{ACSInc}} & 1.0 & \multirow{5}{*}{male} & \multirow{5}{*}{female} & \multirow{5}{*}{44K} 
        			& \multirow{5}{*}{GBM} & $78.98\pm0.58$\% & $78.71\pm0.76$\% &
        			61.04\% & 99.16\% & 10\% & 98\%  \\
        			
        			& & 0.9 & & & & & 
        			$78.49\pm0.75$\% & $79.80\pm0.78$\% &
        			10.56\% & 90.44\% & 0\% & 80\%  \\
        			
        			& & 0.8 & & & & & 
        			$77.53\pm0.75$\% & $80.31\pm0.62$\% &
        			0.23\% & 70.80\% & 0\% & 41\%  \\
        			
        			& & 0.7 & & & & & 
        			$76.71\pm0.70$\% & $80.25\pm0.59$\% &
        			0.05\% & 59.01\% & 0\% & 24\%  \\
        			
        			& & 0.6 & & & & & 
        			$75.93\pm0.77$\% & $80.19\pm0.62$\% &
        			0\% & 50.46\% & 0\%  & 15\%  \\
        			\cmidrule(lr){1-13}
                    \multirow{3}{*}{\makecell{Gender \\ recog.}} & \multirow{3}{*}{\makecell{\texttt{CelebA} \& \\ \texttt{UTKFace}}} &  
                    \multirow{3}{*}{-} &
                    \multirow{3}{*}{young} & \multirow{3}{*}{old} & \multirow{3}{*}{13K.} & 
                    \multirow{3}{*}{CNN} & \multirow{3}{*}{$69.49\pm23.78$\%} & \multirow{3}{*}{$72.39\pm23.79$\%} &
                    \multirow{3}{*}{37.42\%} & \multirow{3}{*}{38.96\%} & \multirow{3}{*}{0\%}  & \multirow{3}{*}{8\%}  \\ \\ \\

        			\bottomrule
		\end{tabular}
	}
	\label{tab:fairness_summary}
\end{table*}

\subsection{Statistical Power}
The first experiments evaluate the statistical power when the goal of the auditor is to detect a target model whose training set underrepresents the $\D_1$ group. Unless otherwise stated, the normative distribution $\D$ is a 50-50 mix of $\D_0$ and $\D_1$, and $\D'=\D_0$.

The results of these experiments are given in the 1--3 and 8th rows of \cref{tab:fairness_summary}.
The table shows that, in the \biased setting, the target model may exhibit higher accuracy and lower mean squared error on $\D_0$ compared to $\D_1$, resulting in a disparate performance across groups.
This was not the case in the control setting, and the training accuracy was significantly higher than the test accuracy, suggesting that the models tend to overfit to their training sets.
This overfitting is consistent with the theoretical analysis in \cref{sec:fairnessaudit}, and as a result the audit using the attack model is successful.



\subsection{Effect Size}
The above experiments involved target models that include \emph{no} training examples from the underrepresented group.
However, this assumption is not necessarily realistic because in practice we would expect \emph{some} examples from a group even when it is underrepresented.
To capture this use case, we now modify the alternative distribution $\D'$.
As before, the normative distribution $\D$ is still a 50-50 mix of $\D_0$ and $\D_1$, but now only an $\beta$ fraction of $\D'$ is $\D_0$, and the rest is distributed as $\D_1$.
We follow the same experimental procedure as before with varying values of $\beta$ for the medical dosing and income prediction tasks to evaluate how the auditor's performance varies with $\beta$.

Rows 3--12 of \cref{tab:fairness_summary} show that the audit becomes less successful as $\beta$ approaches 0.5.
This is to be expected because, when $\beta=0.5$, the target model in the \biased setting follows the normative distribution, leaving the auditor with no means to distinguish the control setting from the \biased setting.
Indeed, when $\beta=0.6$, the auditor's performance is not significantly different from random guessing, but when $\beta=0.9$, she has a much higher likelihood of correctly detecting the GDS.
Therefore, if the auditor audits a model and concludes that there is a GDS, this conclusion is likely to be indicative of a relatively large deviation from the normative distribution.\looseness-1

\subsection{Comparison to a Naive Audit}
We now compare our audit to a naive audit that simply claims that the training set is unrepresentative if the accuracy disparity between groups is sufficiently large.
To evaluate the naive audit, we measure the difference in target model performance for the $\D_0$ and $\D_1$ groups in the control and \biased settings and use it to compute the naive audit's AUC-ROC.

\cref{tab:fairness_summary} shows the AUC-ROC and TPR($P_{90\%}$) of this baseline audit.
These results compare unfavorably with those of the audit, indicating that this naive audit is insufficient for detecting group shifts except in trivial cases.
In addition, for lower values of $\beta$, the sub-50\% AUC shows that the naive audit is \emph{more likely to be wrong than right}.
This is because, as \cref{tab:fairness_summary} shows, the target model tends to be more accurate on the $\D_1$ group than the $\D_0$ group.
In the first row of \cref{tab:fairness_summary}, the naive audit is successful only because the accuracy disparity is in the same direction as the underrepresentation.
Moreover, if the auditor relies solely on accuracy disparity, she can be misled by other sources of accuracy disparity, such as the learning algorithm~\cite{hooker2021moving} and features~\cite{chen2018why}.

\subsection{Presence of Other Shifts}
Finally, we run an experiment to verify that our auditor specifically detects GDS, i.e., that it does not detect other forms of DS.
For this experiment, we induce a shift from $\D$ to $\D'$ without changing the marginal distribution of the protected attribute.
In particular, $\D$ is a 50-50 mix of young and old faces, drawn from the \texttt{CelebA} dataset.
$\D'$ is still a 50-50 mix of young and old faces, but these faces are drawn from a combined \texttt{CelebA} and \texttt{UTKFace} dataset.

The final row of \cref{tab:fairness_summary} contains the results of this experiment.
The target model has an unusually high variance in the \biased setting---this is because the target model sometimes fails to learn patterns that apply to both \texttt{CelebA} and \texttt{UTKFace}.
When this happens, the attack model attains near-perfect accuracy on both young and old faces.
However, this results in no inter-group difference in the attack model's performance with respect to age, so we do not consider this an evidence of underrepresentation.
As a result, the auditor's true positive rate is indistinguishable from random guessing, suggesting that the GDS audit specializes in detecting \emph{group} distribution shifts.

\section{Related Work}\label{sec:related}
In the literature on ML security, we find several attacks that allow to recover information about a model's training set. Ateniese et al.\ first proposed property inference attacks against~\cite{ateniese2015hacking} and, follow-up work, has adapted them to neural networks to infer the demographic attribute of members of the training set~\cite{ganju2018property}.\looseness-1

However, these works assume knowledge of the weights of the neural networks they target.~\cite{ganju2018property}.
As we argue in Section~\ref{sec:statement}, it is unrealistic to assume white-box access in an audit without the model holder's collaboration.
Therefore, these techniques are not directly applicable to auditing.\looseness-1

Another attack that allows to learn information about the training set is membership inference. Membership inference allows an adversary with only query access to a model to learn whether a given data point was in the model's training set.
Yeom et al.~\cite{yeom2018privacy,yeom2020overfitting} formally define membership inference and study the role that overfitting of the target model plays on the attack's success.
Our formal definition of the GDS audit is based on their definition; we modify it to convey our goal of learning information about the entire training distribution rather than a specific point.


A recent study by Maini et al.~\cite{maini2020dataset} presents black-box methods to verify dataset ownership.
Dataset inference is related to membership and attribute inference in that it checks the model for knowledge about specific data points.
By contrast, our audit checks for differences between training data distributions.



More recently, Suri and Evans proposed a formalization of DS inference that is similar to \cref{def:midistauditgame}~\cite{suri2021formalizing}. Our definition differs from theirs in that we let the challenger choose between only two distributions $D$ and $D'$ while they consider a set of distribution transformations on $D$. Because we restrict the space of distributions, the advantage of the adversary we consider is an upper bound of the advantage of the adversary in their definition.\looseness-1

Suri and Evans also propose black-box distribution inference attacks and compare them to Ganju et al.'s white-box property inference attacks. The baselines in our experiments are comparable to their black-box based attacks, as they also set a threshold on the performance of the target model. By contrast, we present a more sophisticated attack based on shadow training that relaxes the white-box assumption of property inference attacks. In addition, we further examine the attack's practicality to audit performance disparities across demographics.\looseness-1

\section{Conclusion, Discussion, and Limitations}~\label{sec:discussion}

This paper proposes a new attack that allows an adversary to infer GDS with fewer assumptions than existing property inference attacks. We study the attack's practicality to audit deployed models for disparate performance across demographic groups.
Our evaluation shows that the techniques that we propose are successful in a wide range of real-world scenarios.
This technique can be used by journalists and researchers to detect underrepresentation of minority groups in proprietary datasets, and thus can explain the accuracy disparities observed in commercial models.

A GDS audit may reveal a flawed data collection methodology that overlooks one of the groups and results in an accuracy disparity during deployment.
We argue that detecting and acknowledging the cause of the disparity is the first step in any mitigation strategy.
Audits that are not able to attribute the causes can fail to show that the origin of the bias is the model holder's responsibility.
For example, in the light of an accuracy disparity, the model holder could dismiss the audit by claiming that solving the task for one of the groups is inherently harder.
A finding of a GDS points to specific actions the model holder could take to mitigate the issue (e.g., reexamine data collection practices).

We now give some caveats and elaborate on the audit's practical challenges.

\paragraph{Detection of an audit.}
\looseness=-1
The model holder cannot reliably detect an audit.
The number of queries required by the audit is not abnormally large, and these queries are drawn from the public normative distribution $\D$.
Therefore, it is expected that other use cases will also involve querying according to $\D$.


\paragraph{Audit precision.}
Our results show that the audit has high recall (TPR), i.e., it is successful in detecting a shift given that there is a shift.
On the other hand, we do not measure the audit's precision, i.e., the likelihood of a shift given a positive finding of the audit.
A realistic measurement of precision requires an estimation of how likely a shift is in practice, i.e., the prior probability of a shift.
It is out of the scope of this paper to measure the prevalence of GDS in commercial models.\looseness-1

\paragraph{Implications of a positive GDS.}
Underrepresentation is neither necessary nor sufficient for an accuracy disparity.
The \texttt{Warfar} 80\% experiment suggests that we can detect underrepresentation even when there is no accuracy disparity.
Although such situations may appear to be harmless, the audit's success indicates that the target model's behavior is different across groups, which reflects some form of bias even if there is no accuracy disparity (e.g., in the confidence scores).\looseness-1

\paragraph{Assuming overfitting.}
Although \cref{sec:fairnessaudit} assumes that models are more accurate near training points, this assumption often holds in practice.
For example, deep neural networks can memorize the entire training set and fit random labels~\cite{zhang2017understanding}, so we would expect them to satisfy such assumption.
In addition, we have evaluated the effect of the learning algorithm $A$ to the success of the attacks in distinguishing a DS. From this evaluation, we have identified properties of $A$ (e.g., its learning capacity) that may make the audited model more susceptible to the attack (see \cref{sec:learning_algo}).\looseness-1


\section{Acknowledgments}
This work has been supported in part by NSF Awards \#1943584, \#1916153, and \#1956435.

{\footnotesize \bibliography{main}}

\clearpage
\appendix


\section{The Complex Attack}\label{sec:appendix_complex}
Apart from the attack model that we used to obtain the main results of the study, we have experimented with a more complex model that we describe in this section.

Similarly to the main model (that we also call the Simple model), the Complex attack model receives the input $x$ to the target or shadow model $h$, as well as the output $h(x)$.
When the number of queries $n_q$ or $n_t$ is greater than 1, the attack model first evaluates each query individually, arriving at a real-valued score for each one.
Finally, the sum of these scores is sent through a sigmoid activation layer to reach a single probabilistic prediction.
To evaluate the accuracy of this model, we use a threshold of 0.5 to binarize this probability.\looseness-1

Our evaluations showed that the Simple attack performs as well or better than the Complex attack in all our experiments. We believe that the Simple attack overperforms the Complex attack because most of the signal to distinguish the target and shadow models stems from their outputs, and feeding the original features as input to the attack adds noise.

\section{Model Architecture}\label{sec:appendix_arch}
In this section, we give the implementation details for our experiments in \cref{sec:eval}.
The writing is organized by the use cases explored in our experiments: gender recognition, medical dosing, and demographic (folktables) tasks.

\subsection{Gender Recognition}
For the \texttt{UTKFace} and \texttt{CelebA} image datasets, we build a convolutional neural network (CNN) to classify faces by gender.
In particular, the inputs to the target and shadow models are first sent through 1--2 convolutional layers with 3$\times$3 kernels and then a max pooling layer.
We continue alternating between 1--2 convolutional layers and a max pooling layer until the input has been sent through four max pooling layers.
We then add a fully connected layer of 128 neurons, followed by a fully connected layer of 1 neuron with sigmoid activation.
Except the final fully connected layer, all layers use ReLU activation.


The target and shadow models were trained for 20 epochs, and the attack models for 50 epochs, both with a batch size of 32.
All experiments were run with 50 different random seeds using Python 3.8.3 and TensorFlow 2.3.1 on a Titan RTX GPU.

\subsection{Medical Dosing}
For the \texttt{Warfar} dataset, the target and shadow models consist of a fully-connected layer of 32 neurons with ReLU activation, followed by a fully-connected layer of 1 neuron.
Because this is a regression task, we do not use an activation function after the final layer.


The target and shadow models were trained for 100 epochs, and the attack models for 50 epochs, both with a batch size of 32.
All experiments were run with 50 different random seeds using Python 3.8.3 and TensorFlow 2.3.1 on a Titan RTX or Titan X (Pascal) GPU.

\subsection{Folktables tasks}
For the income and public coverage prediction tasks defined on the US Census data, we follow Maini et al.~\cite{maini2020dataset}: we use a GBM, the learning algorithm that showed the best performance in their evaluation, and use the exact same hyperparameters, as we also use the same dataset.

All experiments were run with 100 different random seeds using Python 3.8.3 and TensorFlow 2.3.1.




\section{Available Data}\label{sec:available_data}


In \cref{def:midistauditgame}, we assume that the auditor knows the size of the target model's training set.
Consequently, in all the experiments we have ensured that the training sets of all models have similar sizes.
Now we challenge this assumption and study the setting where the auditor has limited amount of data with which to train and test the models under her control.
Less data may result in higher variance in the accuracy of the attack model, which in turn may lead to a higher false positive rate for the audit.
To measure the extent to which this is the case, we have repeated the gender recognition experiment in \cref{tab:fairness_summary} for power-of-two fractions of the initial dataset sampled from $\D$.

\begin{table}[ht]
\rowcolors{5}{}{lightgray}
	\centering
	\caption{The attack model's mean accuracy in the control and \biased settings for the gender recognition model, followed by the AUC, and the TPR with the 90th percentile threshold.}
	\resizebox{\columnwidth}{!}{
		\begin{tabular}{ccccc}
\toprule
\multicolumn{1}{c}{} & \multicolumn{2}{c}{\textbf{Attack Accuracy}} & \multicolumn{2}{c}{\textbf{Audit Perf.}}\\\cmidrule(lr){2-3}\cmidrule(lr){4-5}
$\textbf{Fraction}$ & \textbf{Control} & \textbf{Shifted} & \textbf{AUC} & \textbf{TPR (}$\boldsymbol{P_{90\%}}$\textbf{)} \\
\cmidrule(lr){1-5}
                              $1/2^0$ &     51.51±3.82\% &     98.44±1.10\% &        100\% &                                           100\% \\
                              $1/2^1$ &     50.43±1.67\% &     97.67±1.77\% &        100\% &                                           100\% \\
                              $1/2^2$ &     50.55±2.56\% &     96.86±2.10\% &        100\% &                                           100\% \\
                              $1/2^3$ &     52.93±9.06\% &     93.66±4.77\% &      98.41\% &                                           100\% \\
                              $1/2^4$ &    59.16±16.71\% &    74.79±12.79\% &      79.01\% &                                             8\% \\
                              $1/2^5$ &     50.17±1.67\% &    62.33±21.27\% &      64.17\% &                                            36\% \\
\bottomrule

\end{tabular}

	}
	\label{tab:available_data_summary}
\end{table}

\cref{tab:available_data_summary} shows that the gap in mean attack accuracy between control and \biased settings decreases with smaller fractions of the data.
As expected, the variance in both control and biased accuracy increases with less data.
Yet, even for a $1/8$ fraction of the original data, the difference between mean control and \biased accuracy is statistically significant, showing that the auditor can be successful with up to 8 times less data for the gender recognition task.
Therefore, the assumption of knowing the exact size of the target model's training set can be safely relaxed in practice.

\section{Learning Algorithm}
\label{sec:learning_algo}
The learning algorithm $A$ used to train the target model has an impact on the audit's performance.
If $A$ yields different models depending on the training sets, it can boost the attack model's accuracy.
Conversely, if $A$ yields similar models for training sets coming from different distributions, it will be less vulnerable to the attack.
We now test different learning algorithms and identify algorithmic properties that play a role in the audit.

The learning algorithms that we tested are: a CART Decision Tree (\texttt{dt}) with a maximum depth of five levels;
a Logistic Regressor with L2 regularization and a limited-memory BFGS solver (\texttt{logit});
a Support Vector Machine with an RBF kernel (\texttt{svm});
a Gaussian Na\"ive Bayes (\texttt{nb});
a Random Forest with 50 estimators (\texttt{rf}); a Multi-Layer Perceptron with the Adam optimizer (\texttt{nn}); and the Gradient Boosting Machine (\texttt{gbm}) with the hyperparameters in \cite{ding2021retiring}.
We use the implementations provided by the Sklearn library, and unless otherwise stated, we use their default hyperparameters.\looseness-1

\begin{figure}[ht]
    \centering
	\includegraphics[scale=0.56]{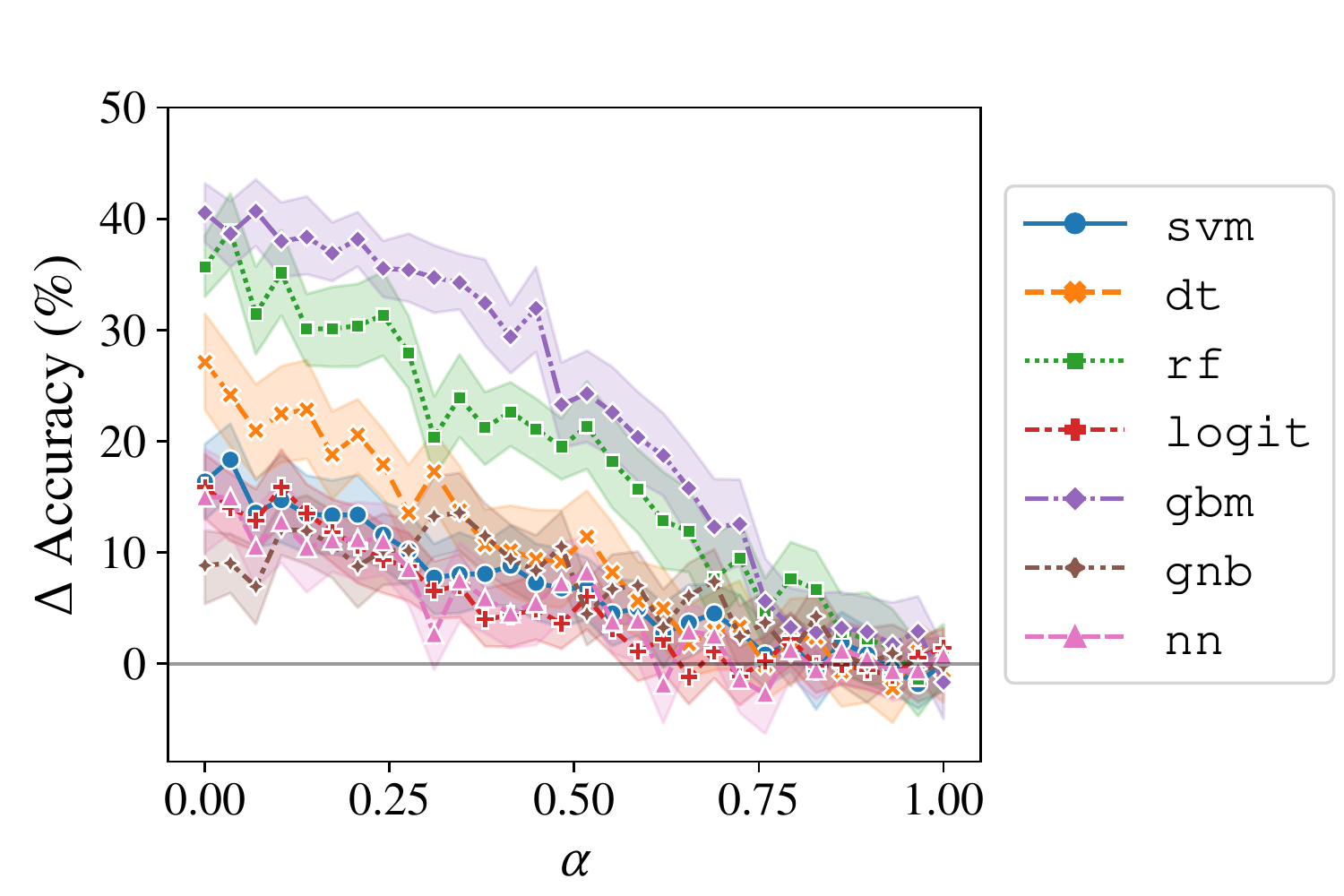}
    \captionof{figure}{$\Delta$Accuracy is the difference in attack accuracy between the control and \biased settings. $\alpha$ parametrizes the mixture distribution. The error bands are the 95\% confidence intervals.}
    \label{fig:linear_comb}
\end{figure}

\begin{table}[ht]
    \centering
    \rowcolors{5}{}{lightgray}
    \captionof{table}{Effect of training the target model with different learning algorithms on the audit. We report the attack model's mean accuracy (and std deviation) in the control and \biased settings for the DS across states, followed by the AUC, and the TPR with the 90th percentile threshold.}
    \resizebox{1.02\columnwidth}{!}{
    \begin{tabular}{ccccc}
\toprule
\multicolumn{1}{c}{} & \multicolumn{2}{c}{\textbf{Attack Accuracy}} & \multicolumn{2}{c}{\textbf{Audit Perf.}}\\\cmidrule(lr){2-3}\cmidrule(lr){4-5}
\multirow{2}*{$\boldsymbol{$A$}$} & \multirow{2}*{\textbf{Control}} & \multirow{2}*{\textbf{Shifted}} & \multirow{2}*{\textbf{AUC}} &  \textbf{TPR} \\
& & & & ($\boldsymbol{P_{90\%}}$) \\
\cmidrule(lr){1-5}
     \texttt{dt} &     53.24±8.77\% &    80.34±12.70\% &      96.68\% &                                            92\% \\
    \texttt{gbm} &     56.26±8.94\% &      96.8±4.43\% &        100\% &                                           100\% \\
    \texttt{gnb} &     54.32±5.60\% &    63.16±11.04\% &      80.94\% &                                            66\% \\
  \texttt{logit} &     53.44±5.57\% &     69.32±8.62\% &      93.38\% &                                            84\% \\
     \texttt{nn} &    53.62±10.69\% &      68.6±9.56\% &      86.80\% &                                            72\% \\
     \texttt{rf} &        53±8.46\% &     88.68±5.78\% &        100\% &                                           100\% \\
    \texttt{svm} &     54.86±6.40\% &    71.24±10.03\% &      91.56\% &                                            78\% \\
\bottomrule
\end{tabular}}
    \label{tab:dos_summary}
\end{table}

\cref{tab:dos_summary} shows the results of the income prediction experiment with a shift along states (California vs.\ Michigan).
To simulate a gradually increasing shift, we sample from the original distributions, denoted by $\D_*$ and $\D'_*$, in proportion to $\alpha\in [0, 1]$: $\D' = \D_*,\;\,\text{ and }\;\,\D = \alpha \, \D_*+ (1-\alpha)\, \D'_*$.

\cref{fig:linear_comb} plots the accuracy of the attack for various values of $\alpha$.
The attack model's accuracy in the control setting is around 0.5 for all $\alpha$.
Thus, we again observe from the figure that the accuracy in the \biased setting is approximately linear in $\alpha$ for all $A$.
We also see that the accuracy depends heavily on $A$ as well.
In particular, the accuracy is substantially lower for \texttt{nn}, \texttt{logit}, and \texttt{gnb} than the other algorithms.
Yet, \cref{tab:dos_summary} shows that the audit's test has high statistical power even for these algorithms.
This means that the attack model's accuracy is sufficient to detect a general DS.
However, this is not the case for higher values of $\alpha$.
At $\alpha=0.75$, the auditor is reduced to the baseline.

\paragraph{Learning capacity.} The audit's lower performance when the target model is trained with \texttt{logit} and \texttt{gnb} can be explained by the relatively low capacity of these algorithms.
When we train the target model on the California dataset (\texttt{IncomeCA}) and test on the Michigan dataset (\texttt{IncomeMI}), the \texttt{gnb} and \texttt{logit} models have some of the lowest training accuracies while having also the lowest standard deviations (see \cref{tab:convex_results_generr}), supporting the hypothesis that these algorithms are producing similar, simple models for both distributions and, thus, decreasing the accuracy of the attack. In comparison, \texttt{dt}, \texttt{gbm}, and \texttt{rf} seem to train complex models with significant overfitting (see \cref{tab:convex_results_generr}). 

\begin{table}[ht]
\rowcolors{1}{}{lightgray}
	\centering
	\caption{Generalization error of the target model for different learning algorithms in the convex combination experiment for $\alpha=0$ in the income prediction task with a shift along sex.}
	\resizebox{0.64\columnwidth}{!}{
	\begin{tabular}{ccc}
\toprule
\multicolumn{1}{c}{} & \multicolumn{2}{c}{\textbf{Accuracy}}\\\cmidrule{2-3}
$\boldsymbol{A}$ & \textbf{Target Train} & \textbf{Target Test} \\
\midrule
     \texttt{dt} &          78.53±0.57\% &         79.25±0.98\% \\
    \texttt{gbm} &           79.7±0.43\% &         80.49±0.83\% \\
  \texttt{logit} &          77.83±0.34\% &         78.82±0.80\% \\
     \texttt{gnb} &          75.85±0.25\% &         76.19±0.75\% \\
     \texttt{nn} &          79.78±0.40\% &         80.07±0.81\% \\
     \texttt{rf} &          99.92±0.02\% &         80.89±0.67\% \\
    \texttt{svm} &           80.6±0.32\% &         80.21±0.78\% \\
\bottomrule
\end{tabular}}
	\label{tab:convex_results_generr_sex}
\end{table}

\begin{table}[ht]
\rowcolors{1}{}{lightgray}
	\centering
	\caption{Generalization error of the target model for different learning algorithms in the convex combination experiment for $\alpha=0$ in the income prediction task with a shift along states.}
	\resizebox{0.64\columnwidth}{!}{
	\begin{tabular}{ccc}
\toprule
\multicolumn{1}{c}{} & \multicolumn{2}{c}{\textbf{Accuracy}}\\\cmidrule{2-3}
$\boldsymbol{A}$ & \textbf{Target Train} & \textbf{Target Test} \\
\midrule
     \texttt{dt} &          79.12±0.38\% &         75.49±1.61\% \\
    \texttt{gbm} &          73.01±1.06\% &         66.78±0.80\% \\
  \texttt{logit} &          78.77±0.31\% &         77.96±0.67\% \\
     \texttt{gnb} &          73.84±0.40\% &         78.06±0.75\% \\
     \texttt{nn} &          79.95±2.01\% &         78.85±2.38\% \\
     \texttt{rf} &          99.87±0.03\% &         76.25±0.83\% \\
    \texttt{svm} &          80.85±0.28\% &         78.95±0.75\% \\
\bottomrule
\end{tabular}}
	\label{tab:convex_results_generr}
\end{table}

\paragraph{Variance.}
We also highlight the \texttt{nn} row in \cref{tab:dos_summary}, which shows that the \texttt{nn} algorithm results in the highest standard deviation of the attack model's accuracy in the control setting.
A closer look at individual accuracy reveals that the \texttt{nn} failed to converge for a few seeds.
When this happens, the attack model can easily distinguish between the target and shadow models, increasing both the mean and standard deviation of the attack model's accuracy in the control setting.

We also observe high variance of the attack's accuracy in the \biased setting for some algorithms.
An algorithm's sensitivity to small changes in the training set can be a source of the variance.
More sensitive algorithms can hinder the auditor, as the increased variance in the attack's accuracy distributions can potentially reduce the audit's statistical power.

\begin{table}[H]
\rowcolors{5}{}{lightgray}
	\centering
	\caption{Effect of various target model learning algorithms on the audit. We report the attack model's mean accuracy (and standard deviation) in the control and \biased settings for the income prediction experiment with the shift along the sex attribute, followed by the AUC, and the TPR in the 90th percentile of the control scores.}
	\resizebox{\columnwidth}{!}{
		\begin{tabular}{ccccc}
\toprule
\multicolumn{1}{c}{} & \multicolumn{2}{c}{\textbf{Attack Accuracy}} & \multicolumn{2}{c}{\textbf{Audit Perf.}}\\\cmidrule(lr){2-3}\cmidrule(lr){4-5}
$\boldsymbol{A}$ & \textbf{Control} & \textbf{Shifted} & \textbf{AUC} & \textbf{TPR (}$\boldsymbol{P_{90\%}}$\textbf{)} \\
\cmidrule(lr){1-5}
     \texttt{dt} &     55.72±8.45\% &    53.86±10.93\% &      43.00\% &                                            10\% \\
    \texttt{gbm} &     57.88±9.92\% &     75.6±11.59\% &      88.70\% &                                            68\% \\
    \texttt{gnb} &      54.1±4.97\% &     54.44±5.16\% &      48.82\% &                                             8\% \\
  \texttt{logit} &      54.3±5.58\% &     60.92±9.96\% &      73.34\% &                                            52\% \\
     \texttt{nn} &     55.32±7.23\% &     62.42±9.64\% &      70.62\% &                                            40\% \\
     \texttt{rf} &     52.82±7.91\% &     60.64±9.85\% &      73.02\% &                                            50\% \\
    \texttt{svm} &     54.22±7.46\% &     62.96±9.80\% &      75.74\% &                                            40\% \\
\bottomrule
\end{tabular}

	}
	\label{tab:probe_summary}
\end{table}

\paragraph{Data characteristics.}
We have also run all the learning algorithms on the income prediction data with the artificial shift along the sex attribute (the results are in \cref{tab:probe_summary}).
Although the audit is overall less effective on these data because the effect is smaller, we observe that \texttt{gnb} and \texttt{gbm} are still the algorithms where the audits achieve worst and best performance. However, the trend is reversed for \texttt{logit} and \texttt{dt}, compared to the results in \cref{tab:dos_summary}.
A look at the models' generalization error (\cref{tab:convex_results_generr_sex}) shows that it is also reversed: now \texttt{dt} does not overfit, but \texttt{logit} does.
Thus, the interplay between the algorithm and the data also plays a role in the audit's success.






\section{Licenses, Privacy, and Informed Consent}\label{sec:priv_consent}

\subsection{US Census dataset (Folktables)}
The data downloaded via the Folktables Python package is from the American Community Survey (ACS) Public Use Microdata Sample (PUMS) files managed by the US Census Bureau. The data itself is governed by the terms of use provided by the Census Bureau\footnote{For more information, see \url{https://www.census.gov/data/developers/about/terms-of-service.html}}.

On that website, it states ``The Census Bureau has created these data to exclude information that would directly identify respondents and characteristics that may lead to the identification of respondents,`` which means that all Personally Identifiable Information (PII) has been stripped from the data.

Participation in the US Census American Community Survey is, by law, mandatory to all US residents and, therefore, consent requirements do not apply.

\subsection{Warfarin dataset}
The International Warfarin Pharmacogenetics Consortium (IWPC) is an international network of 21 research labs that spans over four continents.
IWPC collected the dataset that we used in our \texttt{Warfar} experiments and the data was curated by staff at the Pharmacogenetics and Pharmacogenomics Knowledge Base (PharmGKB) and by members of the IWPC.
Although the dataset is publicly available for download from PharmGKB's website (\url{https://www.pharmgkb.org/downloads}), PharmGKB does not explicitly mention the license that covers the use of this dataset. Thus, we have asked IWPC for permission to use the dataset in our study.

The cohort whose data we use in our study was selected by members of the IWPC for a scientific study~\cite{iwpc}. The authors of this study mention that ``[the] requirement for informed consent was waived because consent had been obtained previously by each participating center, and only de-identified data was used in the study.''

\subsection{The faces dataset}
We use the UTKFace and CelebA datasets. The UTKFace dataset was collected by Zhang, Song and Qi~\cite{utkface} and it is available for download from \url{https://susanqq.github.io/UTKFace}. The CelebA dataset was collected by Liu et al.\ and publised on \url{https://mmlab.ie.cuhk.edu.hk/projects/CelebA.html}. Both datasets are available for non-commercial research purposes only.

All personally identifiable information has been stripped from the tabular data associated to the images. Informed consent is not applicable to these datasets, as the images were collected from the Internet.

\end{document}